\documentclass[sigconf]{acmart}

\usepackage{booktabs} 

\setcopyright{rightsretained}

\acmConference[submitted to SENSYS 2017]{ACM SenSys conference}{November 2017}{Delft, Netherlands} 

\begin{document}
\title{New Directions: Wireless Robotic Materials}


\author{Nikolaus Correll}
\orcid{0002-1911-9277}
\affiliation{%
  \institution{University of Colorado at Boulder}
  \streetaddress{1111 Engineering Drive}
  \city{Boulder} 
  \state{Colorado} 
  \postcode{80309}
}
\email{ncorrell@colorado.edu}

\author{Prabal Dutta}
\affiliation{%
  \institution{EECS Department}
  \streetaddress{University of California, Berkeley}
  \city{Berkeley} 
  \state{California} 
  \postcode{94720}
}
\email{prabal@berkeley.edu}

\author{Richard Han}
\affiliation{%
  \institution{University of Colorado at Boulder}
  \streetaddress{1111 Engineering Drive}
  \city{Boulder} 
  \state{Colorado} 
  \postcode{80309}
}
\email{richard.han@colorado.edu}

\author{Kristofer Pister}
\affiliation{%
  \institution{EECS Department}
  \streetaddress{University of California, Berkeley}
  \city{Berkeley} 
  \state{California} 
  \postcode{94720}
}
\email{pister@eecs.berkeley.edu}

\renewcommand{\shortauthors}{Correll, Dutta, Han and Pister}

\begin{abstract}
We describe opportunities and challenges with wireless robotic materials. Robotic materials are multi-functional composites that tightly integrate sensing, actuation, computation and communication to create smart composites that can sense their environment and change their physical properties in an arbitrary programmable manner. Computation and communication in such materials are based on miniature, possibly wireless, devices that are scattered in the material and interface with sensors and actuators inside the material. Whereas routing and processing of information within the material build upon results from the field of sensor networks, robotic materials are pushing the limits of sensor networks in both size (down to the order of microns) and numbers of devices (up to the order of millions). In order to solve the algorithmic and systems challenges of such an approach, which will involve not only computer scientists, but also roboticists, chemists and material scientists, the community requires a common platform --- much like the ``Mote'' that bootstrapped the widespread adoption of the field of sensor networks --- that is small, provides ample of computation, is equipped with basic networking functionalities, and preferably can be powered wirelessly.
\end{abstract}

\maketitle

\section{Introduction}

\emph{Robotic materials} are an exciting new class of multi-functional composites wherein distributed coordinated computation, sensing, actuation, and communication are embedded within the substrate of a material\cite{mcevoy15}.  Such materials
have the potential to create composites with unprecedented functionality and dynamics. By embedding and abstracting functionality, such materials will revolutionize the creation of autonomous systems. For example, materials that can change their stiffness and shape would allow the creation of aircraft that adapt their aerodynamic profile to varying flight modes, saving energy and reducing noise, while exhibiting the agility of a bat and the efficiency of a goose. Similarly, intelligent robots could be created from materials like smart muscles that can be excited to specific configurations, skin that pre-processes high-bandwidth sensory information, bones that self-repair, eyes that pre-process low-level features, and brain that provides distributed computational power. Ship hulls and building 
facades could be made to actively shield noise, and autonomous cars could be created from tires that can recognize the ground they are rolling on, a
suspension that recuperates energy, a frame that doubles as battery and so on. Such materials can also augment humans, for example by enabling textiles that can sense pose and physiological features of its wearer, fabrics that can dynamically breathe and change permeability/porosity to water and wind depending on conditions, shape-changing and color-changing fabrics, active bandages that can monitor and dynamically seal over a wound, and furniture that automatically adjusts height, shape/form, fit, load-bearing, and function to the user. All of these applications might be enabled by sensing, actuation,  computation and communication that is tightly integrated with polymers.

\begin{figure*}
\includegraphics[trim={0 14cm 0 0},clip,width=0.45\textwidth]{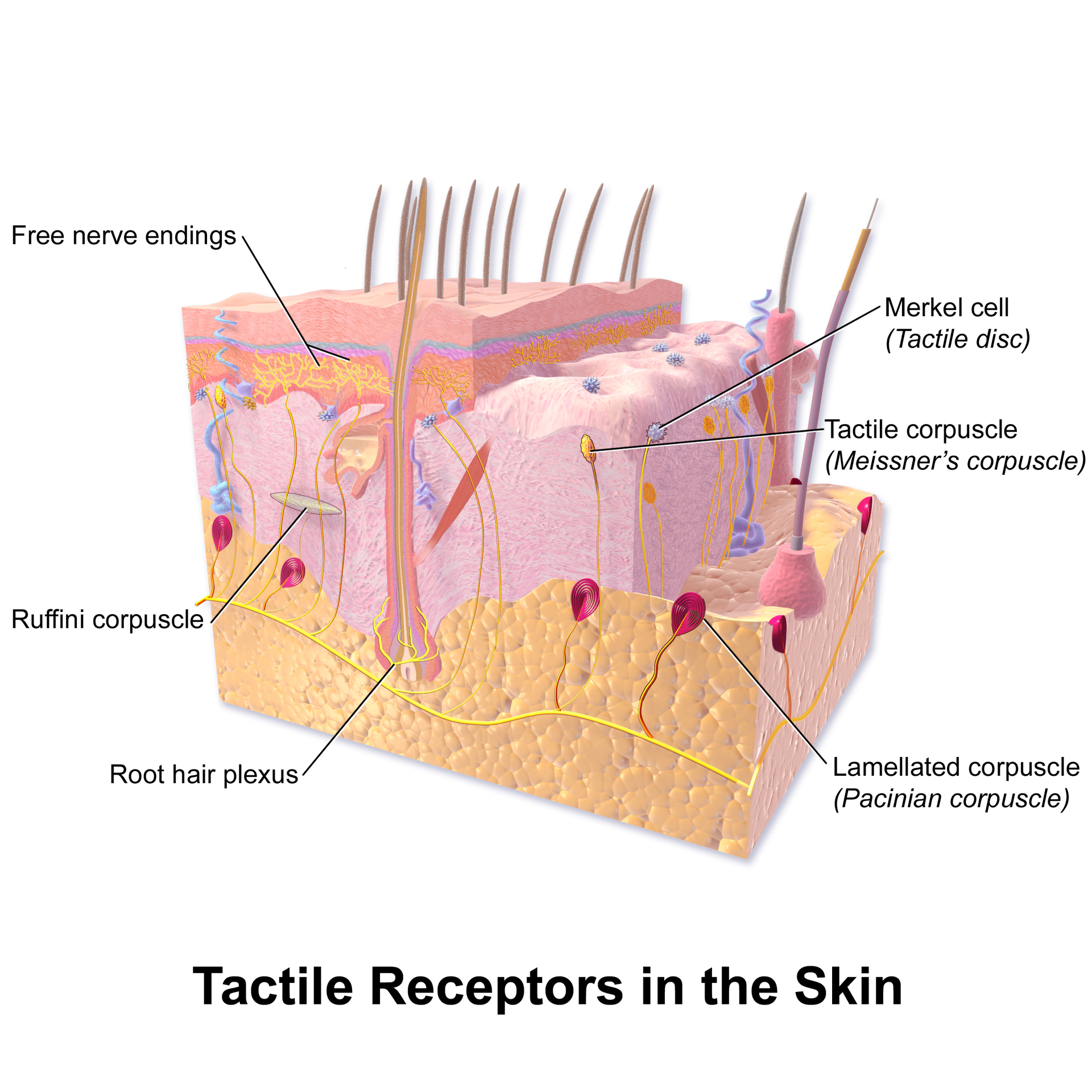}
\includegraphics[trim={0 0.3cm 0 0},clip,width=0.45\textwidth]{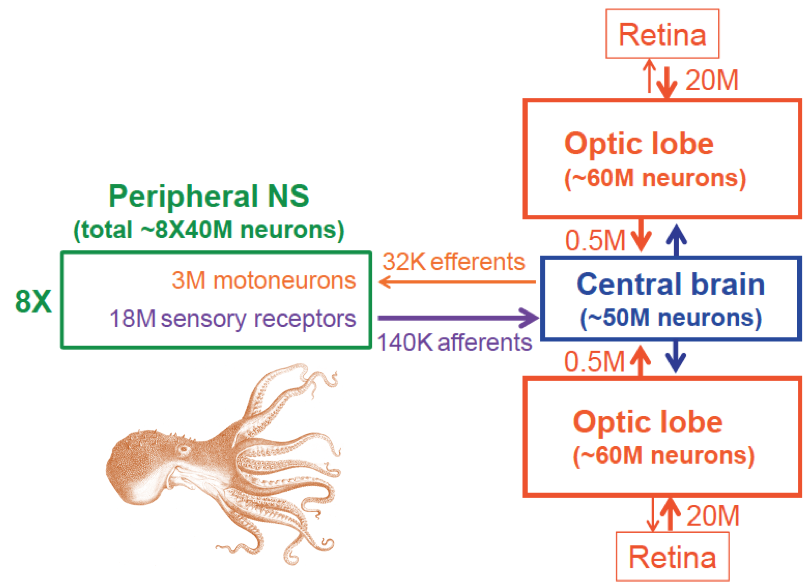}
\caption{Left: Mechanoreceptors in the human skin cover a wide dynamic range, sampling and preprocessing signals up to hundreds of Hertz. From \cite{blausen}, \copyright CC BY 3.0. Right: Distribution of neurons in the Octopus body. Roughly 2/3 of the animal's neurons are located in the arms, where they enable autonomous motion. From \cite{hochner2012embodied}, \copyright Elsevier, reprinted with permission. Both systems suggest integration of micro and meso-scale features into heterogeneous composites. \label{fig:bio} }
\end{figure*}

The recent Workshop on Robotic Materials held at the University of Colorado Boulder March 10-12, 2017 ~\cite{WRM2017} explored the vision and research opportunities for a new generation of robotic materials.  The workshop brought together researchers in sensor networks, robotics, and material science, with keynotes on the biology of octopus motor control and the biology of human skin.  The keynote on the biology of skin highlighted its distributed sensing structure and heterogeneous integration of a variety of sensors, actuators, and support systems, illustrated in Figure \ref{fig:bio}, left. The keynote on octopus motor control highlighted that there are many examples in nature that exhibit decentralized distributed control for locomotion and sensing. Indeed, two-thirds of the Octopus neurons are found in its arms, enabling them with impressive autonomy \cite{hochner2013nervous}, Figure \ref{fig:bio}, right. 

The key outcomes of the discussions at this workshop were that the future of robotic computational materials depends on the tight interdisciplinary collaboration of material scientists, engineers, and computer scientists with expertise in robotics, sensor networks, novel sensors, actuators, and their integration into structural materials. In particular, it became clear that wiring within the materials need to be minimized for manufacturability, complexity and structural properties. Finally, the workshop has led to a preliminary definition of a robotic material as a composite material in which, if you cut it in half at some scale, is still functionally a robotic material.  This definition emphasizes that a key property of a robotic material is that it's composed of many small computing, sensing, actuation and communication elements distributed throughout the material. This integration does not need to be made at the nano-scale, but can also happen at the micro- or centimeter scale, much like the biological examples shown in Figure \ref{fig:bio}, which often consists of heterogeneous materials with micro- and centimeterscale features. 

\subsection{Related Work}
The idea of smart materials being functionalized by miniature wireless devices sprinkled throughout has been first articulated in \cite{berlin1997distributed} and has been driven by the advent of micro-electro mechanical systems (MEMS). Summarizing the results from a DARPA-sponsored workshop, these ideas have led to the concepts of ``smart dust'' \cite{warneke2001smart}, ``amorphous computing'' \cite{abelson2000amorphous}, and ultimately laid the foundation for the field of sensor networks \cite{akyildiz2002wireless}. While ubiquitous cell-phone coverage and cloud-computing have made large-scale distributed computing in sensor networks \cite{duckham2012decentralized} less relevant, miniaturization of wireless computing devices \cite{lee2013modular}, the ability to provide wireless power \cite{pannuto2013m3} and advances in manufacturing \cite{mcevoy15} motivate us to revisit these concepts, synthesize relevant results resulting from previous efforts and articulate new challenges for the field. 

\section{Robotic Materials}
\begin{figure*}
\includegraphics[height=1.82in]{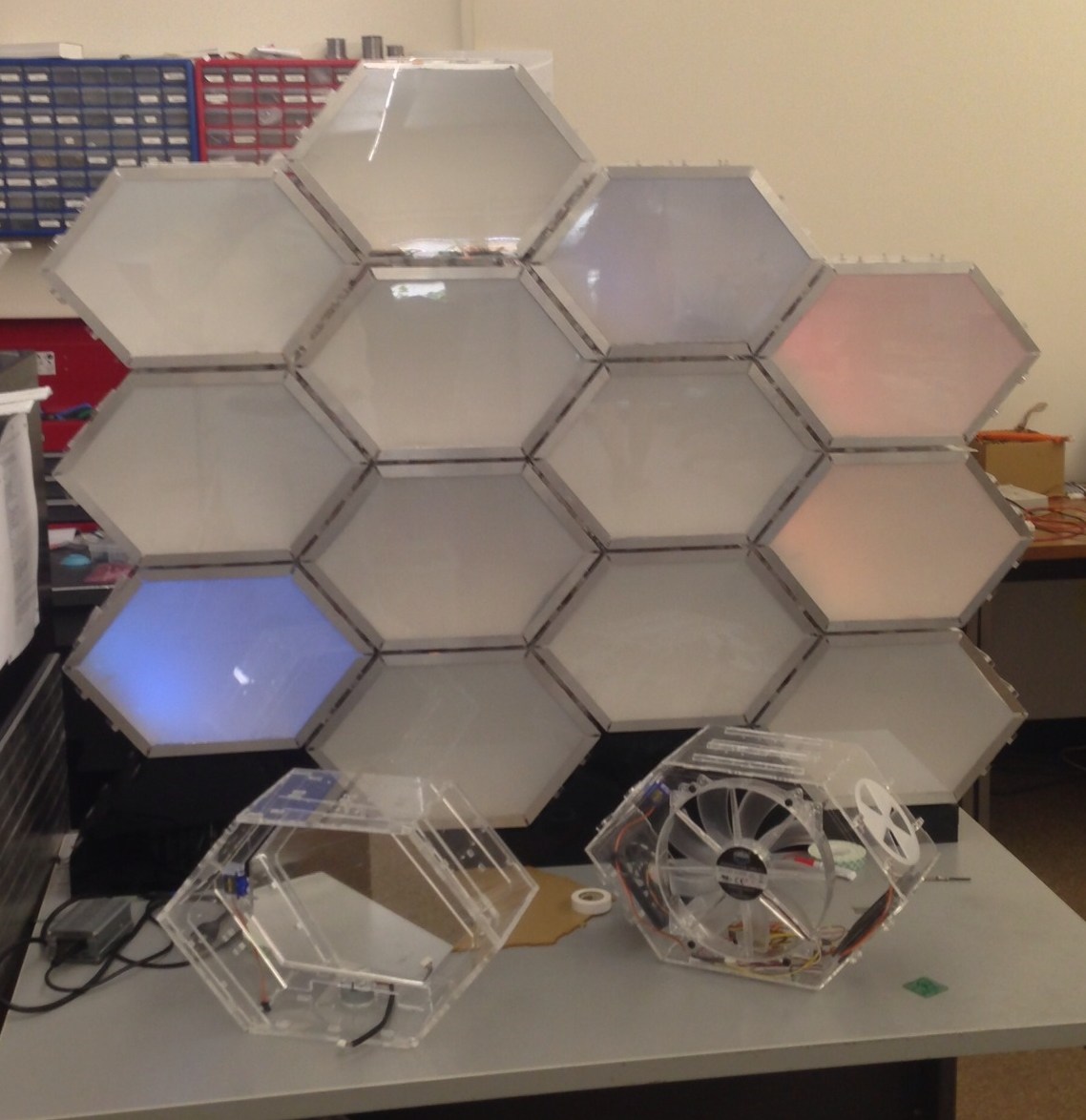}
\includegraphics[height=1.82in,width=2.3in]{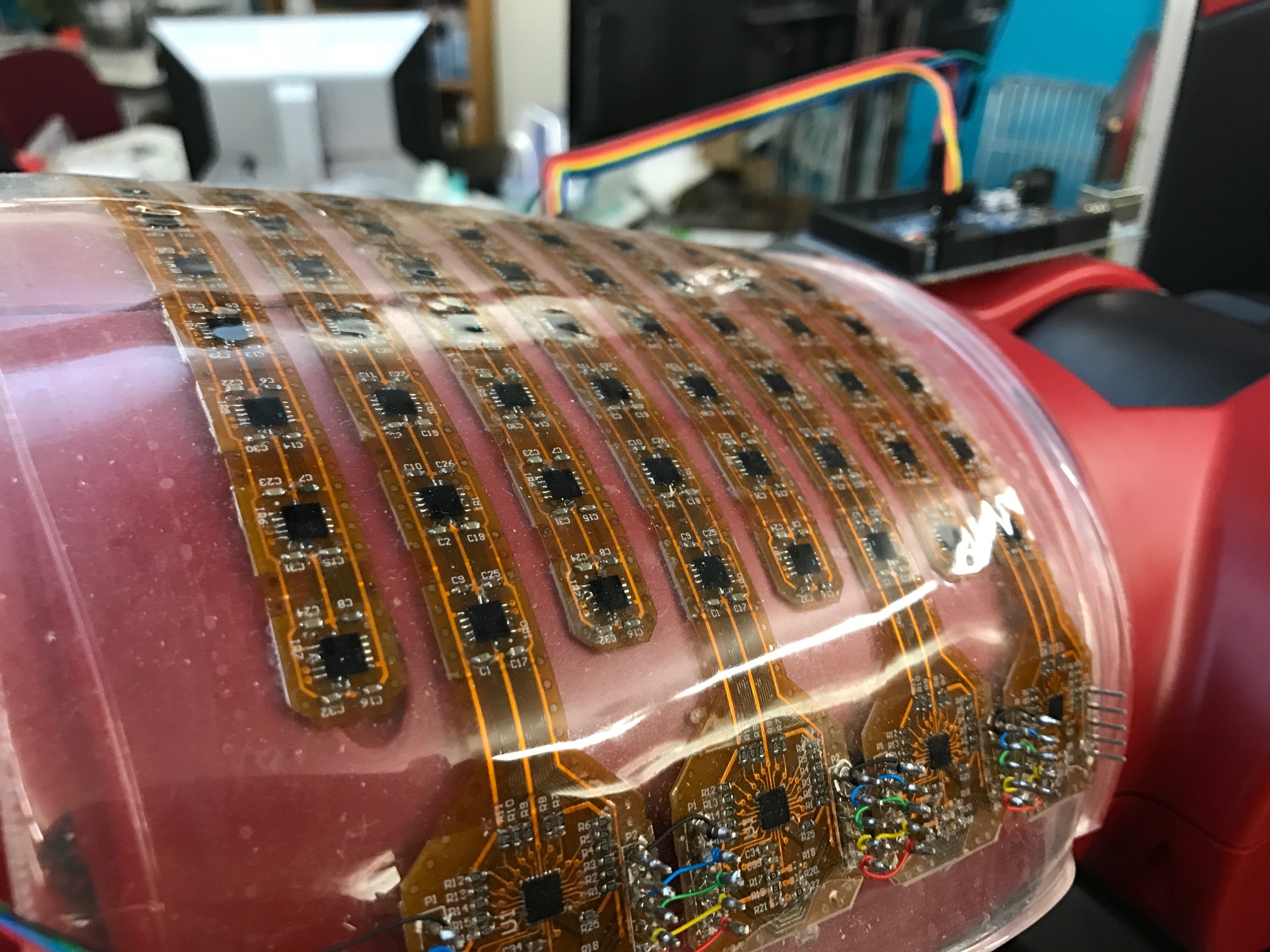}
\includegraphics[height=1.82in]{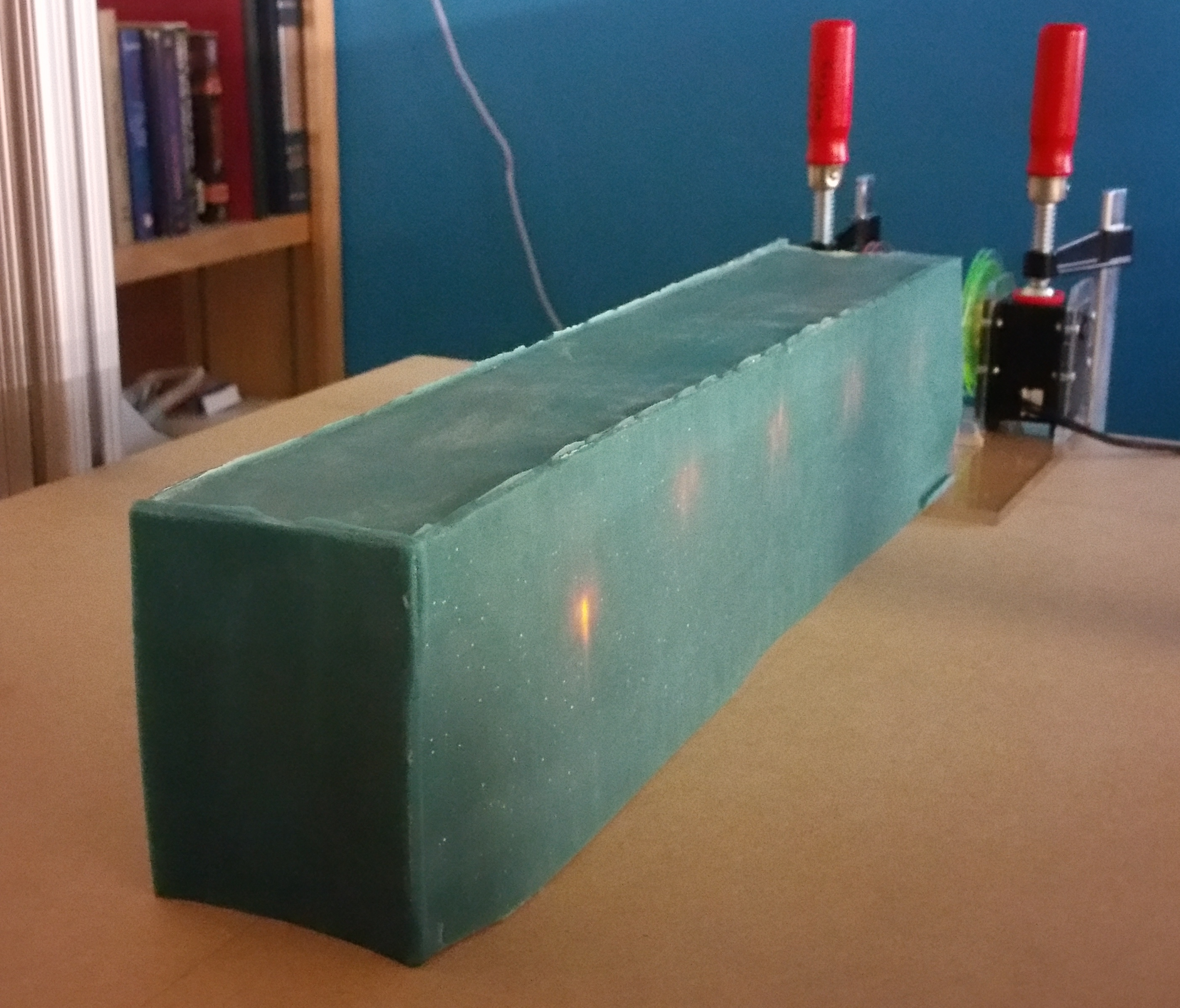}\\
\includegraphics[height=2in]{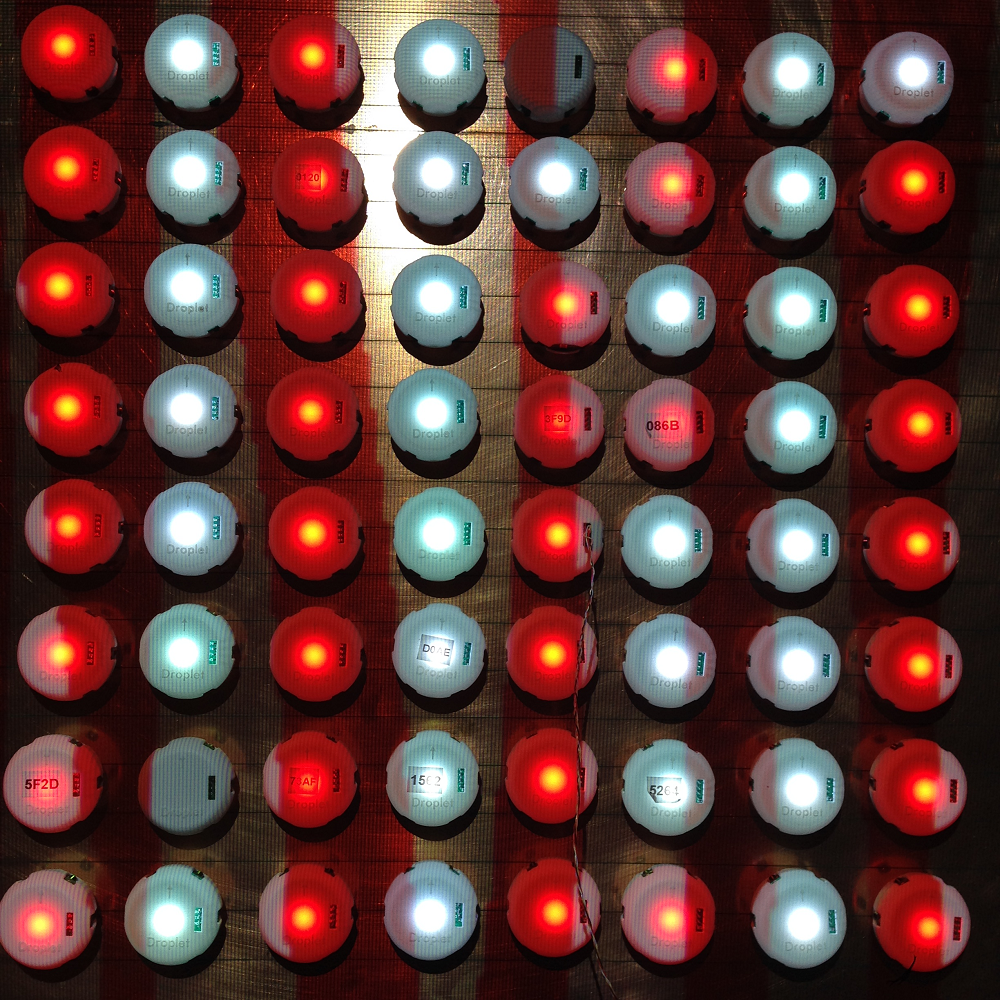}
\includegraphics[height=2in,width=2in]{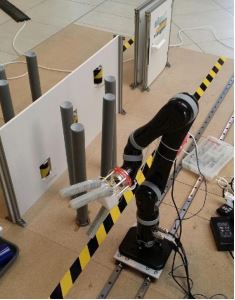}
\includegraphics[height=2in,width=2.2in]{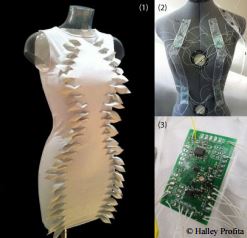}
\caption{From top-left to bottom-right: Amorphous facade \cite{hosseinmardi2015distributed}, tactile sensing skin \cite{hughes2015detecting}, a shape-changing beam \cite{mcevoy2016shape}, active camouflage \cite{li2016}, soft actuators \cite{farrow2016morphological}, and functional wearables \cite{profita2015flutter}.\label{fig:examples} }
\end{figure*}

There exist a series of robotic material prototypes that create novel functionality by tight integration of sensing, actuation, computation and communication, spanning applications in wearables, buildings, and robotics. Figure \ref{fig:examples} shows a series of examples including a facade that can change its opacity and color as well as recognize letters drawn onto it by a user's hand \cite{hosseinmardi2015distributed}, a robotic skin that can recognize social touch \cite{hughes2015detecting}, a beam that can change its shape by calculating its inverse kinematics in a distributed way \cite{mcevoy2016distributed} and varying its stiffness \cite{mcevoy2016shape}, a swarm of robots that can reproduce patterns projected onto it for camouflage \cite{li2016}, soft actuators with integrated curvature sensors that can control their shape and can be arranged into a hand \cite{farrow2016morphological}, and a dress that can localize the direction of incoming sound and display it to its wearer by localized flutter \cite{profita2015flutter}. All the devices shown here demonstrate functionality that would be attractive when realized as a material, but are far from the integration that is necessary to be perceived as such. Yet, all, except for the robotic skin, of which only a single patch with 8x8 sensors is shown, are fully \emph{distributed}, which makes their operation \emph{robust} toward individual failure, and \emph{scalable} toward the number of units. All devices, except the camouflage system, which uses infrared wireless configuration, are wired. Note, that wired communication is the main reason for the arrangements of individual units into regular patterns in these examples, which is not a defining criterion of robotic materials. Figure \ref{fig:examples}, bottom right, also illustrates the challenges with wiring, which make the dress difficult to manufacture and susceptible to failure. 

What would it take to raise the integration level of the above examples, leading to smart glass (facade), smart rubber (skin, beam and fingers), smart paint (camouflage), or smart textiles?
We envision a future in which the application and deployment of robotic materials will be widespread.  We believe such materials will employ wireless communication and power to achieve self-healing adaptive and morphing properties of the material.  Following the vision of wireless sensor networks, we believe such composite materials will consist of many low cost, lightweight and miniaturized elements densely networked together for joint sensing, computing, communication and actuation.

\section{Implementation Issues}
Computation in a robotic material will depend on the sensors and actuators that it integrates. Sensing information and actuator controls needs to be communicated, and the whole system needs to be powered. 

\subsection{Sensing and actuation} 
We envision \emph{sensors} as sparsely distributed as required to recognize human-scale gestures on a facade and as dense as pressure sensors in the human skin. Information generated by sensors in these two examples ranges from binary \cite{hosseinmardi2015distributed} to high-bandwidth signals in the order of hundreds of Hertz \cite{hughes2015texture}, introducing the notions of \emph{bandwidth} in bits per second and \emph{spatial bandwidth} in bits per second per square meter. Other sensors of interest for robotic materials include curvature, radiation, sound, temperature, acceleration, magnetic field, inertia and many others. All of this information needs to be --- to varying degrees --- processed in place or in a restricted area around where the information has been generated, and transferred to a sink. 

We envision \emph{actuators} as sparsely integrated as window panes that can change their opacity (in the order of meters), to materials that can change their stiffness at centimeter resolution and materials integrating thousands of millimeter-sized actuators to realize something like an Octopus arm. Providing control information to each actuator from a centralized location becomes quickly infeasible, be it due to limitations in wiring, addressing, or routing. This is particularly the case if the (spatial) bandwidth requirements of the actuators are high. Analogous to sensing, we envision high-level control to be emitted from a central location, and resolved into regional and local patterns that can be executed from computing elements that are co-located with each actuator. 

\subsection{Wireless communication}
Both sensing and actuation will require \emph{wireless communication} with a dense sensor network of hundreds or thousands of micro-nodes embedded within a small geographic area of a robotic material.  Do our conventional notions of wireless communication with sensor nodes that are fairly separated in distance relative to their communication range break down when considering such dense WSNs?

First, at what radio frequency band do we communicate with these tiny micro nodes?  At least two factors affect the choice of frequency for wireless robotic materials.  Some substrates may absorb certain wavelengths, so the best frequency for wireless communication may be domain specific, e.g., an airplane foil's morphable substrate may exhibit quite different absorption characteristics than an artificial skin containing water, which attenuates microwave frequencies including GHz bands.  A second limiting factor is the length of the antenna itself, which is constrained on micro-nodes.  It is very difficult to make an antenna efficient unless its length is at least one tenth of 
the wavelength of interest.  For example, patch antennas used in today's mobile phones to communicate at low GHz frequencies (10-30cm wavelength) are typically a couple of cm in length.  Physical limitations on the length of such antennas in these micro-nodes of 0.1-1 cm would constrain the frequency bands for effective communication to tens or hundreds of GHz, i.e. mm-wave communications.  The challenge then is to balance these competing factors of material absorption of RF energy with the physical form factors of antennas to choose the appropriate communication band.  

Second, how do we achieve wireless communication with a collection of micro-nodes that are \emph{densely} distributed within the robotic material?  Most deployed wireless sensor networks are relatively sparse collections of nodes in which each transmission interferes with a relatively small number of neighbors.  In a wireless robotic material, transmission by micro-nodes and/or base station APs may jam the communication of large numbers of neighbors, if not the entire material.  Reducing the transmission power to reduce interference is effective in only certain cases because it also reduces the received signal strength.  In certain applications, reducing the transmission power may be acceptable, e.g. the robotic material may be shielded by a surface shell as in a morphable airplane wing that is opaque to external electromagnetic interference, so that the received signal to noise ratio (SNR) is still sufficient to extract the signal.  In other cases, the robotic material may be open to external RF interference, such that the reduced transmission power is swamped by noise, causing an unacceptable SNR.  What other techniques can we apply to deal with node density, such as spread spectrum techniques or hierarchical networking?  

One intriguing technique for reducing interference that we may employ is to equip each micro-node in the robotic material with directional antennas.  This raises numerous interesting research challenges.  On a small scale, how do we implement directional antennas within each micro-node?  That is, how do we design and manufacture low cost electronically steerable phased array antennas within each micro-node for these robotic materials? On a large scale, how do we coordinate directional transmissions across a network of hundreds or thousands of micro-nodes so as to minimize interference?  Further, maintaining directional links requires good localization, so that the sender knows where to point towards the receiver.  If the material is deforming the physical shape of the material, then localization becomes even more challenging.  Given a directionally networked WSN, do conventional sensor networking MAC \cite{polastre2004versatile,buettner2006x} and multi-hop routing protocols \cite{intanagonwiwat2000directed,gnawali2009collection} have to be redesigned to simultaneously accommodate both directionality and scale?  Given the capability to directionally steer communication transmissions to particular locations throughout the material, we may further utilize this feature to directionally beam power to nodes or even directionally heat local regions of the material to change its physical properties, such as its local elasticity, opacity, and permeability.  

Instead of using RF for communication, an intriguing possibility is to employ a network of tiny lasers for communication between micro-nodes and with APs.  In this case the requirements for alignment of lasers and their receivers becomes far more precise and demanding than employing RF.  If the material itself is deforming, then maintaining the link becomes even more challenging. An intermediary solution is to use infrared communication, which is directional, and can also be used to infer range and bearing between individual nodes \cite{farrow2014miniature}. This approach has been used, for example, in \cite{li2016}, shown in Figure \ref{fig:examples}, bottom, left. 

\subsection{Wireless Power}
We envision that \emph{wireless power} is an appealing option for supplying energy to each micro-node.  Each node's antenna would be used to harvest the RF energy, not unlike passive RFID technology.  This raises a number of fascinating research issues.  How do we power the entire ensemble of micro-nodes in the material?  Is the power continuous or intermittent?  Can we enable micro-nodes to have some form of lightweight energy storage?  Should the power be RF-based or would other frequencies such as visible light be better suited for energy harvesting?

First, how do we provide wireless power to every micro-node in the material?  One approach would be to bathe the entire material with a diffuse electromagnetic power source, perhaps at RF or optical wavelengths.  Another approach is to provide targeted directional power to each micro-node.  For example, we could imagine that a charging node or a set of such nodes (within or external to the substrate) knows the precise location of each node in the medium and sends a targeted beam to power each micro-node.  Just as for communication, this would require likely cm accuracy localization in order to place the nodes with sufficient precision within the material.

Second, do we envision that the entire substrate will be powered all the time, or is the power source intermittent?  One scenario is that all micro-nodes are always powered by the charging node(s), so there is no interruption in power, enabling each micro-node to be always on.  In these scenarios, duty cycling becomes unnecessary.  However, we also imagine for example in shape-changing windows and furniture that nodes need not always be powered on, and can be powered on on demand.  Indeed, if micro-nodes are distributed in a uniform array within the material, then selective powering of nodes can be accomplished by having a charging node sweep along each row in raster scan fashion and selectively charging only those nodes needing to become active, much like an electron gun powered pixel phosphors in older cathode ray tubes (CRTs).

Third, unlike passive RFID, for some applications we may permit some lightweight energy storage in each micro-node, allowing it to maintain persistent state.  For most but not all robotic materials, we envision that the weight of chemical batteries precludes them from practical use.  However, it would be interesting to explore whether a lightweight rechargeable capacitor could be integrated into each micro-node.  This would permit each node to retain state even when not being externally charged.  In this case, duty cycling techniques that feature prominently in the WSN literature then become important to preserve energy until the next charging event.

Finally, wireless power raises a number of other key research issues.  What kind of spectra make the most sense for powering at this scale, such as RF, ultrasound, infrared, or optical frequencies?  Can we deliver sufficient power to motivate actuators?  Should the power channel be independent of the communication channel?

\subsection{Distributed Computing and Control} 



Although there has been great progress on the component technologies, much work still remains to be done on the coordination and distributed computing fronts.  Biology distributes computation throughout the body, conjoining sensing, computing, and actuation in complex, hierarchical loops. Here, neural organization in the octopus \cite{hochner2013nervous}, the human retina that implements a tremendous amount of preprocessing of image information \cite{gollisch2008rapid}, or peristaltic motion in the colon \cite{grider1986colonic} are prime examples of autonomous materials with distributed control. 

How to realize sophisticated functions in an energy-efficient manner by distributing them hierarchically but in an engineered way remains a major challenge.  Coupled with realistic communications and computational constraints makes this even more of a research challenge---but one we are now on the verge of being able to explore.  And doing it {\em in silico} makes a great deal of sense from an energy perspective due to CMOS's sub-picojoule energy costs for a variety of common mathematical operations.

In addition to build upon and validate hypotheses from biological systems, wireless robotic materials might become part of large-scale distributed neural networks \cite{hughes2016distributed}. Wireless channels can be modeled as synapses with bandwidth constraints and delay, and implement distributed pattern generation and formation algorithms that can adapt to and learn from their environment.  

Here, emphasis needs to be on algorithms that are scalable to the number of units and robust to failure of individual units. These algorithms will build upon the fields of ``swarm intelligence'', amorphous computing \cite{abelson2000amorphous}, and also control theory and network science, which allows to derive distributed algorithms with provable properties. 

\subsection{Fabrication of Integrated Platforms}
A simplest possible implementation of a robotic material could be the dispersal of miniature, wireless computing elements that can interface attached sensors and actuators in a liquid that can then be cured. This approach lend itself to plastic and rubber materials as well as fiber-glass or carbon composites in which the distribution of computing elements can be amorphous. Such an approach also scales to large areas, which are not feasible to achieve with conventional pick and place mechanisms. 

Once the embedded devices are not microscopic in scale, the interface between materials of varying stiffnesses becomes a challenge. Here, transitions in stiffness need to be implemented in a gradual fashion, which could be achieved by coating sensors, actuators, and computing elements with a material with a stiffness that lies in between the embedded devices and the desired structural material. 

In order to receive wide adoption, the required manufacturing processes and materials need to be accessible, easily available and comprehensively documented. Such an community effort could start with a common hardware platform that can be wirelessly powered and provides basic computation, communication and operating system functionality, and examples of its integration with standard composite material manufacturing techniques.

\section{Discussion}
In the fifteen years that have passed since the first SenSys, practical motes have scaled from matchbox-sized systems with a volume of a dozen cubic centimeters to nearly dust-sized systems of a cubic millimeter in volume.  And along with this dramatic volumetric scaling, we have also witnessed a scaling in idle and active power due to 
transistor scaling and
the use of near-threshold computing, new memory cell designs, low-leakage I/O pads, new interconnect buses, and a host of other innovations, small and large.  As a result of these improvements, today's idle power draws range in the picowatt/nanowatt regime for technology that is being commercialized and will soon be available to the market.  Indeed, much of the recent research agenda has focused on modular and reusable components --- sensors, processors, non-volatile memory, batteries, power management units, radios, and other components that can be stitched together into complete mote systems.

One major benefit of this scaling is that it is now possible to power these tiny devices from energy harvested from the environment \cite{pannuto2013m3}.  And, coupled with mostly wired communications within materials and wireless communications beyond materials, it will soon be possible to modularly integrate sensing, computing, communications, and storage directly into future materials.  With the inclusion of remote programmability, for example optically, entire surfaces composed of robotic materials could reprogrammed {\em in situ}, enabling easy experimentation by the research community.

\section{Conclusion and Outlook}
Robotic Materials offer the potential to create material-like systems that provide functionality that cannot be achieved by material science alone. Advances in miniaturization of computation and communication, novel polymers, and novel manufacturing techniques are making robotic materials possible. Communication and power in robotic materials are likely to be wireless, posing the following challenges to the community:
\begin{enumerate}
\item How do we get to a fundamental mote-like substrate and material that everyone can experiment with and innovate upon?  What does this look like?  Are actuators, sensors, processors, and radios spread uniformly, along with computation throughout the material?  What is the material made of?  Micro-motes?  Is it one robotic material that can be applied in many application domains, or do we need domain-specific robotic materials?

\item How do we wirelessly power these devices (sensors, actuators, radios, processors)?  Are existing systems that can provide power in the order of hundred of microwatts per centimeter square \cite{falkenstein2012low,popovic2013low} sufficient? Does wireless power remain an option once not only sensing, computation and communication but also actuation are required?

\item How do we wirelessly communicate with these devices (sensors, actuators, radios, processors)? Does sharing the same channel that is used for wireless power for communication \cite{zhang2013mimo} scale for large number of devices or do power and communication need to be treated separately? 

\item How do we program large-scale distributed computing substrates? Are event-based, local paradigms such as TinyOS \cite{levis2005tinyos} sufficient, or do we need novel spatial programming concepts \cite{beal2012organizing} to simplify the global-to-local programming problem? 

\item What existing distributed and ``swarm-intelligent'' algorithms for pattern detection and formation can we leverage? How can we generate complex global behavior using predominately local interactions and individually simple local computing?
\end{enumerate}

The field of wireless sensor networks challenged researchers to think outside the box when faced within multi-dimensional challenges of limited resources (energy, cost, bandwidth, memory, computation, storage), wireless communication and in situ field deployment to achieve compelling embedded networked sensing applications.  We believe the dawn of an exciting new age of wireless robotic materials is upon us, driven by technology trends in miniaturization of computing hardware, advances in materials science, and reduced manufacturing costs.  The advent of wireless robotic materials builds on top of WSN accomplishments and challenges WSN researchers to deal with the additional multi-dimensional complexity of integrating actuation, miniaturization, density, and scale into WSNs to achieve compelling applications based on property-morphing composite materials.  We hope this paper has been able to highlight some of the most vital research questions that need to be solved by the sensor networks research community in order make this vision of wireless robotic materials into a compelling reality.

\section*{Acknowledgements}
We are grateful to the participants of the 1st Workshop on Robotic Materials and stimulating discussion. This work has been supported by the Airforce Office of Scientific Research (AFOSR) and the Army Research Office (ARO). We are grateful for this support.  
\bibliographystyle{ACM-Reference-Format}
\bibliography{paper} 

\end{document}